\documentclass[10pt,twocolumn]{article}

\usepackage[preprint]{antintlpaper}

\usepackage[utf8]{inputenc}
\usepackage{inconsolata}
\usepackage{marvosym}
\usepackage{multirow}
\usepackage{adjustbox}

\newcommand{\PAR}{PAR}
\newcommand{\G}{\ensuremath{G}}

\AntTitle{\PAR{}: Progress-Aligned Recurrence for Length Generalization in State Tracking}
\AntRunningTitle{\PAR{}: Progress-Aligned Recurrence for Length Generalization}
\AntAuthors{
Pan Wang\textsuperscript{1*} \and
Yihao Hu\textsuperscript{1*} \and
Guangyuan Dong\textsuperscript{2*} \and
Yitong Lin\textsuperscript{1} \and
Hang Wang\textsuperscript{1} \and
Zimi Liu\textsuperscript{1} \and
Zhihao Wen\textsuperscript{1} \and
Jianshe Li\textsuperscript{1} \and
Jiang-Ming Yang\textsuperscript{1} \and
Wei Wu\textsuperscript{1} \and
Yongqi Tong\textsuperscript{1}\Letter
}
\AntAffiliations{\textsuperscript{1}Ant International \quad \textsuperscript{2}National University of Singapore}
\AntContact{
Contact: Pan Wang \href{mailto:diogenescask@gmail.com}{diogenescask@gmail.com}

$\textrm{\Letter}$ Corresponding Author: Yongqi Tong
\href{mailto:yoqitong@gmail.com}{yoqitong@gmail.com}
}
\AntDate{\today}

\AntKeywords{looped language models, length generalization, state tracking, recurrent depth, latent reasoning}
\AntAbstract{%
Looped language models can run more iterations at test time, but additional iterations need not help them solve longer problems. For sequential state tracking, final-answer supervision does not specify how far each iteration should advance, and repeated updates can change the scale of the recurrent state. We introduce \textbf{Progress-Aligned Recurrence (\PAR{})}, which trains the prediction at iteration $k$ to match the state after the first $k$ operations and rescales recurrent states to their first-iteration norms. On swap-$S_5$, a Qwen2.5-0.5B model trained on at most 32 moves achieves 95.5\% mean full-state accuracy at 512 moves, or $16\times$ the training length, compared with 55.7\% reported for RLT and 0.8\% for explicit chain of thought on the same backbone. Controlled $S_5$ ablations show that removing norm anchoring reduces the extrapolation factor from $16\times$ to below $2.2\times$, and replacing progress-aligned targets with final-answer supervision reduces it to at most $8\times$. A learned completion readout selects when to answer without a supplied stopping depth. When intermediate labels are unavailable, prefix self-distillation induces the correspondence from shorter problem prefixes using only final answers. The gains apply across sequential state-tracking tasks, establishing the task structure required for depth extrapolation. These findings support training intermediate predictions and state scale together to make additional recurrent depth useful for longer problems.
}

\begin{document}

\makeanttitle

{\renewcommand{\thefootnote}{}%
\footnotetext{\textsuperscript{*}Equal Contributions.}}

\section{Introduction}

Length generalisation asks whether a model trained on short problems can solve longer instances of the same task
\citep{anil2022exploring,fan2024looped}. Sequential state tracking provides a controlled setting for studying this
question: each operation updates a state used by later operations \citep{liu2023shortcuts}. A rule-table chain,
for example, uses each retrieved value to select the next lookup, while object tracking updates the locations of
objects after each move \citep{shapiro2026retrofitting,suzgun2023bbh}. In these tasks, the rules remain unchanged as
the sequence grows; the model must apply them correctly over more steps. Throughout this paper, length refers to
the number of task operations.

Looped language models offer a way to make computational depth grow with this length. They repeatedly apply a shared
transformer block to hidden states, allowing additional computation without generating a text description of each
step \citep{dehghani2019universal,geiping2025scaling}, and can be added to pretrained language models
\citep{mcleish2025retrofitted}. Iterations have thus been proposed as a scaling axis alongside parameters, improving
test-time reasoning \citep{geiping2025scaling} and following predictable training trends \citep{prairie2026parcae}.
On longer problems, however, neither axis helps by itself: on $S_5$ trained with at most 16 operations, an untied
model with four times the core depth and a recurrent model run with more iterations both extrapolate less than
$2\times$, and fixed-depth Qwen2.5 models from 1.5B to 7B stay near chance at $16\times$ (\S\ref{sec:main},
\S\ref{sec:scale}). Prior work demonstrates length generalisation by tying the loop count to task length
\citep{fan2024looped}, and shows that intermediate-state supervision can teach a pretrained model to perform
successive rule lookups \citep{shapiro2026retrofitting}. We build on these results to study how to sustain accurate
execution over much longer operation sequences.

A central training choice is what to ask each intermediate loop to predict. A final-answer target does not specify
the state after only some operations. On graph reachability, \citet{chen2026thinking} finds that asking for the final
answer at every loop encourages early predictions and weakens extrapolation. Intermediate-state targets define progress
more directly, but their effectiveness also depends on the recurrence itself. In our state-tracking experiments,
such targets without norm anchoring lose accuracy beyond short extrapolation distances (\S\ref{sec:ablation}).
We therefore study progress-aligned supervision together with control of the scale of states passed between loops.

\begin{figure*}[t]
\centering
\includegraphics[width=\textwidth]{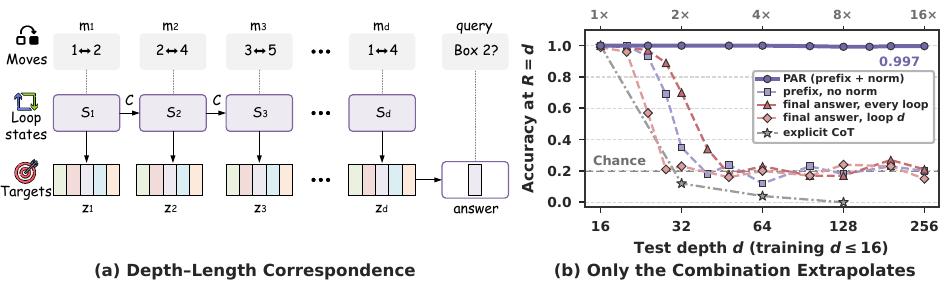}
\caption{(a) \textbf{Depth--length correspondence}: iteration $k$ predicts intermediate task state $z_k$, scaling compute with problem length. (b) Test accuracy on single-query $S_5$ (trained on depth $\le16$, chance 0.2, dashed); \PAR{} sustains near-perfect accuracy to $16\times$ ($d=256$), whereas prior supervision recipes collapse (\S\ref{sec:ablation}).}
\label{fig:teaser}
\end{figure*}

We introduce \textbf{Progress-Aligned Recurrence (\PAR{})} for pretrained language models. At a supervised iteration
$k$, the prediction is trained to match the state after the first $k$ operations (Figure~\ref{fig:teaser}a).
This \emph{depth--length correspondence} assigns successive task stages to successive loop outputs: on a rule-table
chain, for instance, iteration $k$ predicts the value reached after $k$ lookups. To control state scale, \PAR{}
rescales each token's recurrent state to its first-iteration root-mean-square norm. The targets specify the intended
progress, while norm anchoring fixes state magnitude as the shared block is reused.

The correspondence also guides when to read the answer and how to obtain intermediate targets. A correct answer can
deteriorate if the model continues looping after completion (\S\ref{sec:halting}). We train the existing output layer
to predict whether execution is complete, and use a short confirmation window to select the answer without a supplied
stopping depth. When intermediate labels are unavailable, we use prefixes that are themselves valid shorter
problems. Our \emph{prefix self-distillation} uses predictions on such prefixes as targets for the
corresponding iterations on the full input, learning from final-answer labels alone.

Our contributions are:
\begin{itemize}[leftmargin=*]
  \item We formulate length generalization in looped models as establishing a depth--length correspondence, identifying state drift and unaligned loop targets as twin failure modes and proposing \textbf{Progress-Aligned Recurrence (\PAR{})} to couple progress supervision with RMS norm anchoring.
  \item We extend \PAR{} with a parameter-free learned halting rule to stop inference autonomously and prefix self-distillation to induce the correspondence without intermediate step labels.
  \item We demonstrate $16\times$ extrapolation on swap-$S_5$ ($95.5\%$ full-state accuracy vs.\ $55.7\%$ for RLT and $0.8\%$ for explicit CoT), show consistent benefits across backbones up to 7B, and isolate the necessity of both components through controlled ablations.
\end{itemize}

\section{Related Work}

\noindent\textbf{Looped and recurrent-depth transformers.}
Weight-sharing across depth originated in Universal Transformers \citep{dehghani2019universal}, and recurrence can extend computation to harder instances \citep{schwarzschild2021learn,bansal2022endtoend}. In language models, recurrent depth has been trained from scratch \citep{geiping2025scaling} or retrofitted into pretrained models \citep{mcleish2025retrofitted}. Looped architectures can simulate chain of thought \citep{saunshi2025reasoning}, exhibit stable scaling behavior \citep{prairie2026parcae}, share representations across depths \citep{jeddi2026loopformer}, and relate to implicit models \citep{schone2025implicit}, though depth is frequently underutilized \citep{dau2026beyond}. We address this by defining what each iteration computes.

\noindent\textbf{Length extrapolation and halting.}
Prior work links loop count to input length \citep{fan2024looped} or explicit reasoning steps \citep{fan2026bridging}. From-scratch looped models examine stochastic stopping \citep{kuo2026stabilizing}, mechanistic limits \citep{liang2026horizon}, stability conditions \citep{labovich2026stability}, and fast--slow recurrence \citep{takashiro2026fastslow}; among them, RLT \citep{zhang2026rlt} achieves strong swap-$S_5$ performance via token-level feedback. In retrofitted models, per-loop supervision yields modest extrapolation \citep{shapiro2026retrofitting}, whereas step-wise final-answer supervision impairs compositional generalization \citep{chen2026thinking}. For terminating recurrence, adaptive computation time \citep{graves2016adaptive} and PonderNet \citep{banino2021pondernet} train halting probabilities with dedicated modules and ponder penalties. Under our correspondence, the existing output head detects completion directly, requiring neither extra parameters nor ponder losses.

\noindent\textbf{Latent reasoning and self-supervision.}
Coconut \citep{hao2024coconut} and stepwise internalization \citep{deng2024explicit} internalize reasoning along the sequence dimension using chain-of-thought data. In contrast, we compute along depth, using intermediate labels or prefix self-distillation inspired by mean-teacher consistency \citep{tarvainen2017mean}, in the spirit of work that derives training signals from a model's own outputs, interactions, or environments \citep{xiong2026dive,wang2026reflex,wang2026diagnosis,hu2026seal,zhao2026towards,tong2026grounded}. While randomized positional encodings improve sequence generalization \citep{ruoss2023randomized}, our positional stretch stabilizes the recurrent step map when scaling to longer contexts without step labels.

\section{Depth--Length Correspondence}
\label{sec:method}

\begin{figure*}[t]
\centering
\includegraphics[width=\textwidth]{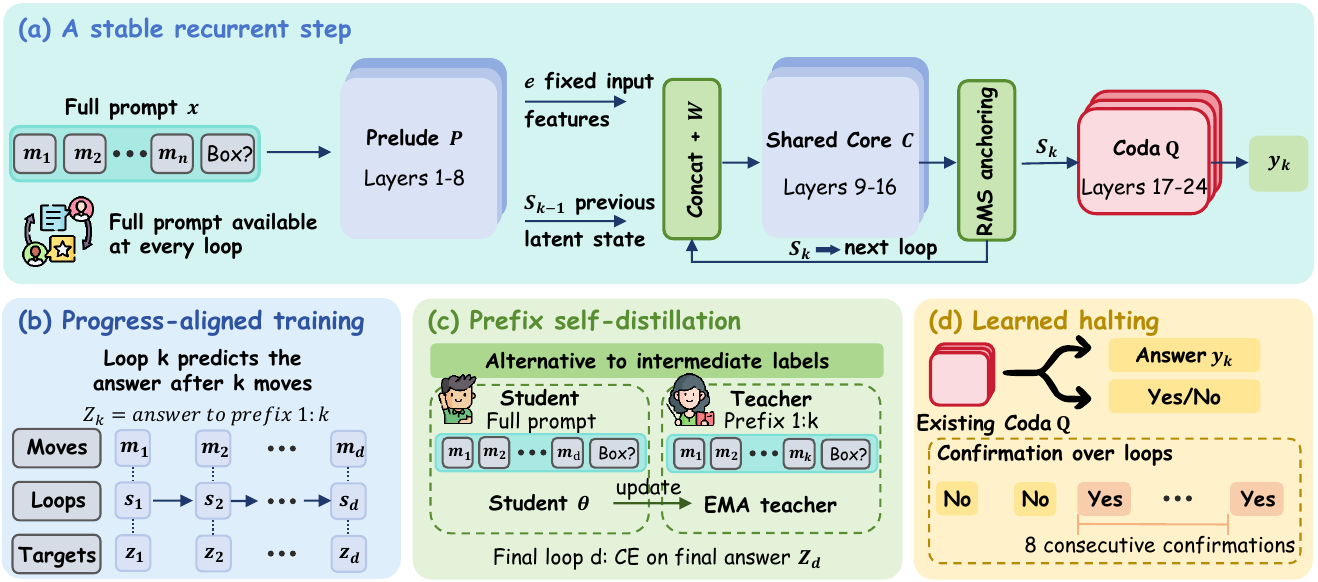}
\caption{Overview of the \PAR{} framework.
\textbf{(a) Architecture (\S\ref{sec:model}, \S\ref{sec:stationary}):} Retrofitted prelude $P$, weight-tied core $C$, and coda $Q$, stabilized across iterations via RMS norm anchoring.
\textbf{(b) Progress alignment (\S\ref{sec:targets}):} Recurrent iteration $k$ predicts intermediate state $z_k$ after $k$ operations, establishing an explicit depth--length correspondence.
\textbf{(c) Prefix self-distillation (\S\ref{sec:sdmethod}):} An EMA teacher evaluated on prefix $1{:}k$ supervises iteration $k$, inducing intermediate states from final-answer labels alone.
\textbf{(d) Learned halting (\S\ref{sec:haltmethod}):} The coda jointly predicts task completion (\texttt{Yes}/\texttt{No}); inference terminates when a multi-step confirmation window is met.}
\label{fig:method}
\end{figure*}

Figure~\ref{fig:method} outlines the \PAR{} framework. We realize the depth--length correspondence by combining retrofitted recurrence with RMS norm anchoring to maintain a stationary step map (\S\ref{sec:model}, \S\ref{sec:stationary}), supervising each loop with progress-aligned targets (\S\ref{sec:targets}), reading completion autonomously via learned halting (\S\ref{sec:haltmethod}), and extending the supervision to label-free settings through prefix self-distillation (\S\ref{sec:sdmethod}).

\subsection{A retrofitted looped language model}
\label{sec:model}
Following \citet{mcleish2025retrofitted}, we split the 24 layers of Qwen2.5-0.5B \citep{qwen2025qwen25} into a
prelude $P$ (layers 1--8), a weight-tied core $C$ (layers 9--16) and a coda $Q$ (layers 17--24), as shown in Figure~\ref{fig:method}a. With static input features $e=P(x)$ and $s_0=e$, the adapter $W$ projects the concatenated representations into the core, iterating
\begin{equation}
  s_k = C\!\left(W[e;\,s_{k-1}]\right),\qquad k=1,\dots,R,
  \label{eq:loop}
\end{equation}
and reads an answer $y_k = Q(s_k)$ from any iteration. The adapter $W\in\mathbb{R}^{h\times2h}$ is initialised to
$[0\,|\,I]$, so with $R=1$ the retrofitted model initially computes the same function as its parent. All weights are
trained, gradients flow through the last 8 iterations, and answers are read with a classification head over the
candidate answer tokens at the query position.

\subsection{Defining an iteration: progress-aligned targets}
\label{sec:targets}
Consider a problem whose answer is computed by $d$ sequential steps (for state tracking, $d$ moves applied to a state).
Let $z_j$ be the answer after the first $j$ steps, so $z_d$ is the final answer. As depicted in Figure~\ref{fig:method}b, the correspondence asks iteration $k$
to hold $z_k$ until the problem is finished. With step size $c$ (we use $c=1$), the training loss is
\begin{equation}
  \mathcal{L}_{\text{PAR}} = \sum_{k\in\mathcal{K}} \operatorname{CE}\!\left(Q(s_k),\; z_{\min(ck,\,d)}\right),
  \label{eq:par}
\end{equation}
where $\mathcal{K}$ contains the iterations inside the gradient window. Each iteration must advance the computation by
exactly $c$ steps, so the iterations needed for a problem grow as $d/c$ by construction. Supervising $z_d$ at every
iteration breaks the correspondence: early iterations are asked for an answer they cannot yet compute, and the loss
rewards shortcuts that do not scale with $d$.

Two depth schedules satisfy Eq.~\eqref{eq:par}. Under \emph{forced depth}, each batch contains a single depth $d$, the
model runs $R=\lceil d/c\rceil$ iterations, and there is no loss after completion. Under \emph{hold}, $R$ is drawn from
a heavy-tailed distribution ($1+\mathrm{Poisson}(\lambda)$ with log-normal $\lambda$, mean 12, cap 48;
\citealp{geiping2025scaling}) and iterations past $d/c$ keep $z_d$. Forced depth is our default (\PAR); \S\ref{sec:halting}
removes its need to know $d$ at test time.

\subsection{Keeping the step map stationary}
\label{sec:stationary}
Extrapolation applies the step map hundreds of times, far beyond the training depths, so anything that drifts with $k$ or
with absolute position eventually changes what the map computes. We remove two such drifts. \emph{State-norm anchoring}
rescales every token's state to the RMS norm of its first-iteration state after each iteration (Figure~\ref{fig:method}a),
$s_k \leftarrow s_k\cdot\operatorname{rms}(s_1)/\operatorname{rms}(s_k)$, with a stop-gradient on the reference; $R=1$
is unchanged. \emph{Positional stretch} addresses long prompts, in which the query lies far from early moves: during a
short fine-tuning stage we multiply position indices by $s\sim U[1,s_{\max}]$, ramping $s_{\max}$ from 1 to 8
\citep[cf.][]{ruoss2023randomized}, and test with ordinary positions. Label-supervised \PAR{} reaches $16\times$ without
the stretch; the label-free variant did not reach $16\times$ until fine-tuned with it (\S\ref{sec:abl-sd}).

\subsection{Reading completion: learned halting}
\label{sec:haltmethod}
Under the correspondence, whether the computation has finished is a property of the iteration index and the problem,
so it can be supervised. As illustrated in Figure~\ref{fig:method}d, the coda read-out that produces the answer is also trained to put probability on the token
\texttt{Yes} when $ck\ge d$ and on \texttt{No} otherwise, using class-balanced cross-entropy with weight 0.5 on every
supervised iteration plus one earlier iteration as an extra negative. No parameters are added, and training never runs
an iteration beyond $d$. At test time we select the first answer iteration $R$ with
$P(\texttt{Yes})>\tau$ such that at least $k$ of the iterations $R,\dots,R+m-1$ also exceed $\tau$. After checking this
confirmation window, we return the prediction made at $R$. We use $\tau=0.5$, $k=6$, $m=8$;
\S\ref{sec:abl-halt} explains how this rule was chosen.

\subsection{Correspondence without step labels: prefix self-distillation}
\label{sec:sdmethod}
Intermediate answers $z_j$ come for free from a synthetic generator but not from most real data. The correspondence
supplies them anyway: the first $j$ steps of a problem form a shorter problem whose answer is $z_j$, and shorter problems
are the ones a model learns first. As shown in Figure~\ref{fig:method}c, prefix self-distillation therefore uses only final answers. Iterations
$k\ge\lceil d/c\rceil$ receive cross-entropy on $z_d$. For each example, up to $n_k$ earlier iterations $k<\lceil d/c\rceil$
are trained to match, by KL divergence, the answer distribution of a teacher that reads the same prompt truncated
after $ck$ steps, decoded at its own completion iteration. The teacher is an exponential moving average (EMA) of the
student \citep{tarvainen2017mean} or the live student.

\subsection{Training}
We use AdamW with learning rate $2{\times}10^{-5}$ ($10^{-4}$ for the adapter), 200 warm-up steps and one epoch over
${\approx}400$k examples spread uniformly over training depths (compute details in Appendix~\ref{app:extra}). No
intermediate tokens are generated at training or test time.

\section{Experimental Setup}
\label{sec:setup}

\noindent\textbf{Tasks.}
\emph{Group word problems.} A prompt lists $d$ moves on five boxes, each holding a distinct item, and queries a box.
In \emph{swap-$S_5$} each move is the identity or one of the 10 transpositions; in $A_5$ each move is a 3-cycle.
Chance accuracy per query is $0.2$. \emph{Swap-$S_5$ protocol} \citep{grazzi2025unlocking,zhang2026rlt}: we
train on at most 32 moves and test on $1{,}024$ sequences per length. A sequence counts as correct only if all five
boxes are answered correctly (\emph{full-state accuracy}). \emph{Benchmarks from prior literature}: the rule-table chain of
\citet{shapiro2026retrofitting} (unary depth, train $\le8$, test to 64) and the graph reachability task of
\citet{chen2026thinking} (train $\le5$ hops, test to 24).

\noindent\textbf{Metrics.}
Accuracy is reported in \% unless stated otherwise. On $S_5$ and $A_5$ we also report the extrapolation factor $\G=d_{50}/d_{\text{train}}$,
where $d_{50}$ is the first test depth at which accuracy with $R=d$ iterations falls below the midpoint between chance
and 100\%. With test depths up to 256 and $d_{\text{train}}=16$, $\G=16$ is the largest measurable value.

\noindent\textbf{Baselines.}
With the same retrofitted Qwen2.5-0.5B backbone, data and step budget as \PAR{}, we compare explicit CoT fine-tuning
that writes the full state trace, a fixed-depth model ($R=1$), an untied model with the core repeated four times
(48 layers), and random-$R$ recurrence with state-norm \citep{geiping2025scaling}. On the swap-$S_5$ protocol we also compare with Transformer and RLT, using the values reported by
\citet{zhang2026rlt} for their best split, $4+4$ (mean over 3 seeds). Variants of \PAR{} that remove one of its components, including the
supervision schemes of prior work, are studied separately in \S\ref{sec:ablations}.

\section{Main Results}

\subsection{Depth extrapolation}
\label{sec:main}

\begin{table*}[t]
\centering
\small
\setlength{\tabcolsep}{4.2pt}
\begin{tabular}{@{}lccccccc@{}}
\toprule
 & \multicolumn{4}{c}{$S_5$, train $d\le16$} & $A_5$, train $d\le16$ & \multicolumn{2}{c}{Swap-$S_5$, train $L\le32$} \\
\cmidrule(lr){2-5}\cmidrule(lr){6-6}\cmidrule(l){7-8}
Method & 32 ($2\times$) & 64 ($4\times$) & 128 ($8\times$) & $\G$ & $\G$ & 256 ($8\times$) & 512 ($16\times$) \\
\midrule
Transformer & -- & -- & -- & -- & -- & $0.85\pm0.30$ & $0.85\pm0.30$ \\
RLT & -- & -- & -- & -- & -- & $97.3\pm2.8$ & $55.7\pm25.8$ \\
Explicit CoT & 12 & 4 & 0 & 1.247 & 1.125 & 0.8 & 0.8 \\
Fixed depth ($R=1$) & 19 & 20 & 17 & 1.62 & 1.32 & -- & -- \\
Untied $4\times$ core (48 layers) & 39 & 14 & -- & 1.76 & -- & -- & -- \\
Random-$R$ recurrence$^*$ & 55 & 20 & 25 & 1.95 & 1.05 & -- & -- \\
\midrule
\PAR{} (ours) & \textbf{100} & \textbf{100} & \textbf{99.3} & \textbf{16} & \textbf{16} & $\mathbf{99.5\pm1.2}$ & $\mathbf{95.5\pm11.0}$ \\
\bottomrule
\end{tabular}
\caption{Main results. Accuracy (\%) with $R=d$ iterations on $S_5$ (one query, chance 20\%) and full-state accuracy
(\%) on the swap-$S_5$ protocol of \citet{zhang2026rlt}; column headers give the test length and its multiple
of the training length. $\G$: extrapolation factor. $^*$Best $R$ per depth.
\PAR{}: mean over 3 seeds on $S_5$, 2 on $A_5$ and 6 on swap-$S_5$ (1{,}024 test sequences per length).
--: not reported.}
\label{tab:main}
\end{table*}

Table~\ref{tab:main} compares \PAR{} with the baselines. On the swap-$S_5$ protocol, \PAR{} maintains over 95\%
full-state accuracy out to $16\times$ extrapolation, where RLT drops to 55.7\% and every \PAR{} seed exceeds RLT's mean.
On $S_5$ and $A_5$, \PAR{} achieves the maximum measurable extrapolation factor ($\G=16$) across all seeds, whereas
fixed-depth, recurrent, and 48-layer untied baselines stay below $2\times$: neither iteration count nor parameter count
alone substitutes for the correspondence. Explicit CoT is perfect within the training range but collapses to chance by
$8\times$ on swap-$S_5$, demonstrating that the pretrained backbone does not inherently confer extrapolation.

\subsection{Larger and different backbones}
\label{sec:scale}

\begin{table}[t]
\centering
\small
\setlength{\tabcolsep}{3pt}
\begin{adjustbox}{max width=\columnwidth}
\begin{tabular}{@{}llccc@{}}
\toprule
Method & Seeds & 1.5B & 3B & 7B \\
\midrule
\multirow{2}{*}{Fixed depth} & mean & $22.2_{\pm3.8}$ & $21.3_{\pm2.4}$ & $18.2_{\pm3.8}$ \\
 & best & 26.5 & 24.0 & 21.5 \\
\midrule
\multirow{2}{*}{\PAR{}} & mean & $\mathbf{71.7}_{\pm42.7}$ & $\mathbf{57.8}_{\pm42.5}$ & $\mathbf{80.7}_{\pm33.1}$ \\
 & best & \textbf{100} & \textbf{100} & \textbf{100} \\
\bottomrule
\end{tabular}
\end{adjustbox}
\caption{Larger Qwen2.5 backbones: accuracy (\%) at $d=256$ ($16\times$) on $S_5$ (train $d\le16$, one query,
chance 20\%). mean: mean$_{\pm\text{sd}}$ over 3 seeds; best: best seed.}
\label{tab:scale}
\end{table}

\begin{table}[t]
\centering
\small
\setlength{\tabcolsep}{3.5pt}
\begin{adjustbox}{max width=\columnwidth}
\begin{tabular}{@{}llcccc@{}}
\toprule
 & & \multicolumn{4}{c}{Test depth $d$} \\
\cmidrule(l){3-6}
Backbone & Method & 32 & 64 & 128 & 256 \\
\midrule
\multirow{2}{*}{Qwen3-0.6B} & Fixed depth & 59.2 & 17.7 & 21.3 & 22.0 \\
 & \PAR{} & \textbf{99.7} & \textbf{87.5} & \textbf{60.2} & \textbf{36.2} \\
\midrule
\multirow{2}{*}{Qwen3-1.7B} & Fixed depth & 50.0 & 19.8 & 20.3 & 19.2 \\
 & \PAR{} & \textbf{99.8} & \textbf{75.5} & \textbf{43.8} & \textbf{38.8} \\
\midrule
\multirow{2}{*}{Qwen3-4B} & Fixed depth & 61.0 & 18.2 & 20.3 & 17.7 \\
 & \PAR{} & \textbf{97.0} & \textbf{62.3} & \textbf{62.3} & \textbf{53.0} \\
\bottomrule
\end{tabular}
\end{adjustbox}
\caption{Qwen3 backbones: accuracy (\%) by test depth ($2\times$--$16\times$ the training depth) on $S_5$ (train
$d\le16$, one query, chance 20\%), mean over 3 seeds.}
\label{tab:qwen3}
\end{table}

Table~\ref{tab:scale} evaluates larger Qwen2.5 backbones (1.5B, 3B, and 7B) on $S_5$. On every backbone, the best \PAR{}
seed maintains 100\% accuracy through $d=256$, whereas fixed-depth baselines remain near chance ($\le26.5\%$). Across seeds,
\PAR{} provides substantial gains at all model sizes, with non-extrapolating seeds accounting for cross-seed variance.

Table~\ref{tab:qwen3} reports results on the Qwen3 family \citep{yang2025qwen3} across 0.6B, 1.7B, and 4B scales. \PAR{}
consistently outperforms fixed depth across all evaluated depths: accuracy remains near ceiling at $2\times$ and maintains
a wide margin through $8\times$. At the $16\times$ horizon ($d=256$), where fixed-depth models collapse near chance,
\PAR{} reaches up to 53\% mean accuracy on 4B, retaining an advantage across scales despite increased seed variance on
larger models.

\subsection{Learning when to stop}
\label{sec:halting}

\begin{table}[t]
\centering
\small
\setlength{\tabcolsep}{3.5pt}
\begin{tabular}{@{}lcc@{}}
\toprule
Method (swap-$S_5$, train $L\le32$) & 256 ($8\times$) & 512 ($16\times$) \\
\midrule
RLT & $97.3\pm2.8$ & $55.7\pm25.8$ \\
{\color{gray}\PAR{}, oracle $R=L$} & {\color{gray}$98.7\pm3.1$} & {\color{gray}$93.2\pm16.3$} \\
\midrule
\PAR{} + learned halting & $\mathbf{98.7\pm3.1}$ & $\mathbf{92.9\pm16.2}$ \\
\bottomrule
\end{tabular}
\caption{Learned halting: full-state accuracy (\%) without access to the problem length $L$ (6 seeds, 1{,}024
sequences per length, step-6000 checkpoints). Grey: the same models run for exactly $L$ iterations, for reference.}
\label{tab:halt}
\end{table}

Forced-depth models answer correctly at $R=L$ but degrade rapidly thereafter, dropping to near-zero accuracy within a
few extra iterations at $16\times$. Inference therefore requires identifying when execution finishes. With the ``6 of 8''
rule (\S\ref{sec:haltmethod}), the jointly trained read-out halts without knowing problem length $L$ (Table~\ref{tab:halt}):
accuracy closely matches the oracle iteration count across all seeds and reaches $92.9\%$ at $16\times$ (vs.\ $55.7\%$ for RLT).

\begin{figure*}[t]
\centering
\includegraphics[width=\textwidth]{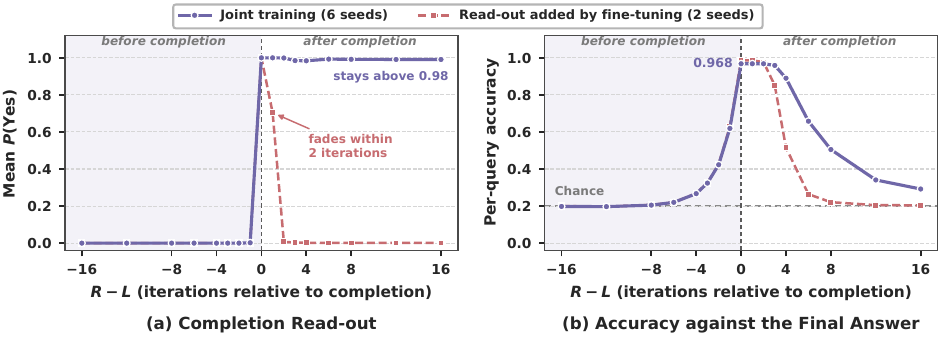}
\caption{Halting at $16\times$ the training length ($L=512$, all 5 queries). Solid purple: joint training (6 seeds,
1{,}024 sequences); dashed red: read-out added by fine-tuning (2 seeds, 1{,}024 sequences). The shaded region marks
iterations before completion ($R<L$) and the dashed vertical line marks $R=L$. (a) The completion read-out
stays at 0 until the iteration count reaches $L$. For jointly trained models it then stays near 1, while for the fine-tuned
read-out it fades within two iterations. (b) Accuracy against the \emph{final} answer peaks exactly when the read-out
fires and decays a few iterations later, so the answer must be read where the ``finished'' signal starts. Before $L$
the iterations hold intermediate states, which differ from the final answer (horizontal dashed line: chance).}
\label{fig:halt}
\end{figure*}

This stopping mechanism succeeds because jointly trained models enter a persistent finished state (Figure~\ref{fig:halt}a),
with $P(\texttt{Yes})$ jumping from near zero before $L$ to near 1 afterwards. Because the answer representation itself remains
valid for only a few steps past completion (Figure~\ref{fig:halt}b), the rule reliably reads the prediction from the onset
of the finished signal.

\subsection{Learning without step labels}
\label{sec:labelfree}

\begin{table}[t]
\centering
\small
\setlength{\tabcolsep}{3.5pt}
\begin{tabular}{@{}lcc@{}}
\toprule
Method (final-answer labels only) & $d{=}256$ ($16\times$) & $\G$ \\
\midrule
Fixed depth ($R=1$) & -- & 1.62 \\
Untied $4\times$ core (48 layers) & -- & 1.76 \\
Random-$R$ recurrence & -- & 1.95 \\
Final answer at every loop & 21 / 21 & 2.14 / 1.99 \\
Final answer at loop $d$ & 15 & 1.48 \\
Final answer at loop $d$, norm & 26 / 19 / 36 & $<$4 / $<$4 / 8 \\
\midrule
\PAR{} + prefix self-distillation & \textbf{85} & $\mathbf{\ge16}$ \\
\bottomrule
\end{tabular}
\caption{Extrapolation without intermediate labels on $S_5$ (train $d\le16$, one query, chance 20\%; accuracy in \%,
``/'' separates seeds). The final-answer recipes are defined in Table~\ref{tab:ablation}. --: not evaluated at
$d=256$; these models are at chance by $d=48$ (Table~\ref{tab:s5depth}).}
\label{tab:sd}
\end{table}

Prefix self-distillation learns the correspondence from final answers alone. Among the label-free models in
Table~\ref{tab:sd}, it is the only approach sustaining high accuracy at $16\times$ (85\% at $d=256$, $\G\ge16$), whereas
alternative final-answer supervision recipes degrade substantially. While label-supervised \PAR{} remains stronger at the
longest depths (Figure~\ref{fig:teaser}b), positional stretch fine-tuning (\S\ref{sec:abl-sd}) enables strong extrapolation on
swap-$S_5$: fine-tuned models reach up to 99.7\% accuracy at $16\times$ (Table~\ref{tab:sdruns}), surpassing RLT without
requiring step labels.

\subsection{Prior benchmarks and scope}
\label{sec:boundary}

\begin{table}[t]
\centering
\small
\setlength{\tabcolsep}{3.5pt}
\begin{tabular}{@{}lcc@{}}
\toprule
Method & Rule chain & Graph reach. \\
 & ($\le8\to64$) & ($\le5\to24$) \\
\midrule
Final answer at every loop & 1.45 & 2.4 \\
Final answer at loop $d$ & -- & 3.3 \\
Prefix targets, no norm & 1.78 & \textbf{3.45} \\
\PAR-hold & $\mathbf{\ge8}$ & fail$^*$ \\
\midrule
\PAR{} & 6.0 & 2.6 \\
\bottomrule
\end{tabular}
\caption{$\G$ on benchmark tasks from \citet{shapiro2026retrofitting} (rule chain) and \citet{chen2026thinking}
(graph reachability). Methods are defined in Table~\ref{tab:ablation}. --: not run. $^*$Predicts the majority class.}
\label{tab:tasks}
\end{table}

\noindent\textbf{Benchmarks from prior literature.}
Table~\ref{tab:tasks} evaluates the recipes on the benchmark tasks on which the final-answer and unanchored recipes
were introduced. On the rule-table chain, an inherently sequential lookup problem, \PAR-hold reaches $\G\ge8$ (94\%
accuracy at depth 64) and \PAR{} reaches 6.0, whereas prefix targets without norm anchoring stall at 1.78.

\noindent\textbf{Inductive bias and task structure.}
Graph reachability is solved by expanding a frontier in parallel rather than by a fixed sequence of steps, so it has
no canonical intermediate state for iteration $k$ to hold; all recipes stay below $3.5\times$ on it. The
correspondence targets sequential multi-step computation, the regime in which fixed-depth transformers fail
\citep{liu2023shortcuts,merrill2024illusion}.

\noindent\textbf{Natural-language format.}
On a BIG-Bench-Hard-style \citep{suzgun2023bbh} \emph{tracking shuffled
objects} format with natural-language swaps, trained on $\le16$ swaps, \PAR-hold with positional stretch reaches
94--96\% at 32 and 79\% at 64 swaps (chance 17\%), whereas random-$R$ and fixed-depth baselines with the same stretch
reach at most 67\% and 21\%. The original BBH items (3, 5 and 7 objects) lie within the training range, and all models
score 95--100\% on them.

\section{Ablation Studies}
\label{sec:ablations}

\subsection{Both conditions are necessary}
\label{sec:ablation}

\begin{table*}[t]
\centering
\small
\begin{adjustbox}{max width=\textwidth}
\begin{tabular}{@{}llcccc@{}}
\toprule
Variant & Target at iteration $k$ & Schedule & Norm & $\G$ & Acc.\ (\%) \\
\midrule
Final answer at every loop & $z_d$ at every $k$ & heavy-tailed $R$ & \checkmark & 2.14 / 1.99 & 21 / 21 \\
Final answer at loop $d$ & $z_d$ at $k=d$ only & forced $R=d$ & -- & 1.48 & 15 \\
Final answer at loop $d$, norm & $z_d$ at $k=d$ only & forced $R=d$ & \checkmark & $<$4 / $<$4 / 8 & 26 / 19 / 36 \\
Prefix targets, no norm, hold & $z_{\min(k,d)}$ & heavy-tailed $R$ & -- & 1.84 & 16 \\
Prefix targets, no norm & $z_{\min(k,d)}$ & forced $R=d$ & -- & 1.81 & 20 \\
\midrule
\PAR-hold & $z_{\min(k,d)}$ & heavy-tailed $R$ & \checkmark & 16 / 12 & 75 / 30 \\
\PAR{} & $z_{\min(k,d)}$ & forced $R=d$ & \checkmark & \textbf{16 / 16 / 16} & \textbf{100 / 100 / 99} \\
\bottomrule
\end{tabular}
\end{adjustbox}
\caption{Ablation of the two conditions on $S_5$ (train $d\le16$, test to 256, one query, chance 20\%). Norm:
state-norm anchoring. $\G$ and accuracy at $d=256$ are given per seed. All rows share the backbone, data, optimiser and
step budget. Rows 1, 2 and 5 instantiate the supervision recipes of
\citet{chen2026thinking}, \citet{fan2024looped} and \citet{shapiro2026retrofitting} on our backbone. Accuracy curves
across test depths with $R=d$ iterations are in Figure~\ref{fig:teaser}b.}
\label{tab:ablation}
\end{table*}

Table~\ref{tab:ablation} varies one condition at a time on $S_5$ with training depth $\le16$. Three findings follow.

\emph{Final-answer targets do not define an iteration.} Supervising $z_d$ at every iteration stalls at $\G\approx2$,
in line with \citet{chen2026thinking}, while supervising it only at iteration $d$ reaches $\G=1.48$. Adding norm
anchoring to iteration-$d$ supervision extends extrapolation for a single seed ($\G=8$), but the remaining seeds degrade
earlier; across seeds, accuracy at depth 256 averages 26.8\%, far below the near-perfect accuracy maintained by \PAR{}.

\emph{The correspondence needs a stationary step map.} Prefix targets without norm anchoring stall at ${\sim}1.8\times$
under either schedule, consistent with findings in unanchored recurrence \citep{shapiro2026retrofitting}, as state
representations drift when iterated repeatedly.

\emph{Together, the two conditions extrapolate at least $16\times$.} \PAR{} reaches $\G=16$ across all seeds with
near-ceiling accuracy at $d=256$. \PAR-hold also extrapolates beyond $12\times$, though with lower consistency across seeds.

\noindent\textbf{Positional stretch alone does not create the correspondence.}
Applied as a fine-tune to the fixed-depth and random-$R$ models, positional stretch leaves both at chance by depth 48
(23\% and 18\%; Table~\ref{tab:s5depth}). Applied to a \PAR-hold model with step size $c=2$, which already has the
correspondence, the same fine-tune keeps 99\% accuracy at depth 512, $32\times$ the training depth, in one run with a
fixed $R=288$. Stationarity makes an existing step map reusable; it does not create one.

\noindent\textbf{A one-pass initialisation blocks the correspondence.}
On $A_5$, \PAR{} trained from the pretrained parent reaches $\G=16$ on both seeds (Table~\ref{tab:main}).
Warm-starting all layers from a fixed-depth model that already solves $d\le16$ in a single pass gives $\G\approx1.4$:
the model keeps its one-pass solution and never learns the step map. Warm-starting only the prelude succeeds on one of
two seeds.

\subsection{Stopping rule and alternatives to halting}
\label{sec:abl-halt}

\begin{table}[t]
\centering
\small
\setlength{\tabcolsep}{3.5pt}
\begin{tabular}{@{}lcc@{}}
\toprule
Stopping rule & 256 ($8\times$) & 512 ($16\times$) \\
\midrule
Single iteration ($k=m=1$) & -- & $77.8\pm25.7$ \\
8 in a row & $\mathbf{98.7\pm3.1}$ & $90.3\pm16.4$ \\
\midrule
6 of 8 (\PAR) & $\mathbf{98.7\pm3.1}$ & $\mathbf{92.9\pm16.2}$ \\
\bottomrule
\end{tabular}
\caption{Stopping rules for the jointly trained read-out (swap-$S_5$, full-state \%, same 6 seeds and checkpoints as
Table~\ref{tab:halt}). --: not reported.}
\label{tab:haltrules}
\end{table}

\noindent\textbf{Stopping rule.}
Evaluating $P(\texttt{Yes})$ at a single iteration is brittle (Table~\ref{tab:haltrules}), as transient probability
spikes cause premature stopping at long horizons. Conversely, requiring all 8 consecutive iterations to confirm
completion is overly strict when probabilities fluctuate, delaying stopping until predictions drift. The ``6 of 8''
rule balances robustness and responsiveness, matching oracle stopping within 1.6 points across all evaluated sequences.

\noindent\textbf{Joint training versus a fine-tuned read-out.}
Adding the read-out by fine-tuning an existing \PAR{} checkpoint yields different dynamics (Figure~\ref{fig:halt}a):
$P(\texttt{Yes})$ spikes sharply at completion and decays within two iterations. As a result, multi-step confirmation
windows fail to trigger, whereas single-iteration stopping recovers oracle accuracy across both seeds
(Table~\ref{tab:haltseeds}). However, fine-tuning causes minor accuracy degradation on one seed at $16\times$ compared to
joint training.

\noindent\textbf{Training on extra iterations is not a substitute.}
An alternative to an explicit halting signal is making completion an absorbing state by training models on extra iterations
that must retain $z_d$. While this preserves in-distribution accuracy, it severely harms extrapolation: accuracy collapses
below 1\% at $16\times$ across evaluated seeds. Asking the recurrent core to advance intermediate states while simultaneously
acting as a fixed point introduces conflicting objectives; an independent completion read-out avoids this competition.

\subsection{Teacher and positional stretch in prefix self-distillation}
\label{sec:abl-sd}

\begin{table}[t]
\centering
\small
\setlength{\tabcolsep}{3.5pt}
\begin{tabular}{@{}lccc@{}}
\toprule
Teacher & $d{=}64$ & $d{=}256$ & $\G$ \\
\midrule
EMA 0.99 + curriculum & -- & -- & 2.6 \\
Slow EMA + confidence filter & -- & -- & 5.6 \\
EMA 0.99, forced depth & \textbf{99} & 57 & ${\approx}16$ \\
EMA 0.99 & 97 & 61 & $\mathbf{\ge16}$ \\
\midrule
Live student + depth curriculum & 98 & \textbf{85} & $\mathbf{\ge16}$ \\
\bottomrule
\end{tabular}
\caption{Teacher configurations for prefix self-distillation on $S_5$ (train $d\le16$, single-query accuracy in \%
at $R=d$). --: not evaluated. The last row is the configuration reported in Table~\ref{tab:sd}.}
\label{tab:teacher}
\end{table}

\noindent\textbf{Teacher.}
Three of five teacher configurations reach $\G\approx16$ or more (Table~\ref{tab:teacher}); the live-student teacher
with a depth curriculum keeps the highest accuracy at $d=256$.

\noindent\textbf{Positional stretch.}
On swap-$S_5$, initial label-free models extrapolate partially to $8\times$ but collapse at $16\times$ (Table~\ref{tab:sdruns}).
The failure is abrupt rather than cumulative: per-query accuracy stays high at 256 moves before dropping to chance at 512,
pointing to a non-stationary step map driven by absolute position encodings. A short fine-tune combining the label-free
objective with positional stretch restores extrapolation across all starting checkpoints, raising $16\times$ accuracy
from near-chance levels up to 99.7\%. Training with stretch from the start similarly yields runs achieving perfect accuracy
at both lengths, while label-supervised \PAR{} achieves its results without stretch.

\section{Conclusion}

A looped language model extrapolates in depth when each of its iterations has a defined job. Aligning iteration $k$
with the first $k$ steps of the problem, and keeping the step map stationary, turns extra iterations into extra
reasoning. A retrofitted 0.5B model trained this way surpasses the best published latent-recurrence model on swap-$S_5$
at $8\times$ and $16\times$ the training length. The same correspondence lets the model learn when to stop and lets it
learn its intermediate states from final answers alone. Defining what an iteration means is thus a training target in
its own right, alongside choosing how many iterations to run.

\section*{Limitations}
We evaluate on algorithmic and templated state-tracking tasks, where intermediate states are well defined; applying
the correspondence to open-ended natural-language reasoning \citep{xiong2025deliberate,xiong2026enhancing,xiong2026adaptive} is left to future work. Tasks without a sequential
decomposition, such as graph reachability, fall outside its scope. Our scaling study covers Qwen2.5 backbones up to
7B and Qwen3 backbones up to 4B, with three seeds each, in the single-query $S_5$ setting only; swap-$S_5$, learned
halting and prefix self-distillation are evaluated only on Qwen2.5-0.5B. On the larger backbones, as with prefix
self-distillation, results vary across seeds far more than for \PAR{} on Qwen2.5-0.5B (standard deviations of
33--43 points at $16\times$ on Qwen2.5-1.5B to 7B), accuracy does not rise monotonically with Qwen2.5 size, and no
Qwen3 backbone approaches Qwen2.5-0.5B at $16\times$. These results therefore do not establish a monotonic trend with
model size; characterizing the mechanisms of seed variance on larger backbones and stabilizing recurrent scaling
remain open questions for future work.

\section*{Ethics Statement}
This work studies training methods on synthetic reasoning tasks with publicly available models
\citep{qwen2025qwen25,yang2025qwen3}. It involves no human subjects and no personal data. Latent reasoning writes no intermediate
steps, which makes it less transparent than chain of thought. Under the correspondence, however, every iteration can be
decoded into the intermediate answer it holds, which offers a way to inspect the latent computation.

\bibliographystyle{plainnat}
\bibliography{custom}

\appendix

\begin{table*}[t]
\centering
\small
\begin{tabular}{@{}lcccccccc@{}}
\toprule
Model & 24 & 32 & 48 & 64 & 128 & 256 & 384 & 512 \\
\midrule
Fixed depth & 0.73 & 0.19 & 0.23 & 0.20 & 0.17 & -- & -- & -- \\
Random-$R$ (best $R$ per depth) & 0.93 & 0.55 & 0.23 & 0.20 & 0.25 & -- & -- & -- \\
Untied 4$\times$ core (48 layers) & 0.88 & 0.39 & 0.20 & 0.14 & -- & -- & -- & -- \\
Fixed + stretch FT & 0.84 & 0.35 & 0.23 & 0.16 & 0.16 & 0.19 & -- & -- \\
Random-$R$ + stretch FT & 0.93 & 0.60 & 0.18 & 0.18 & 0.17 & -- & -- & -- \\
\PAR-hold w/o state-norm & 0.98 & 0.23 & 0.18 & 0.17 & 0.20 & -- & -- & -- \\
\PAR-hold ($c{=}1$), $R{=}d$, seed 0 & 1.00 & 1.00 & 1.00 & 0.97 & 0.95 & 0.86 & 0.72 & 0.72 \\
\PAR-hold ($c{=}1$), $R{=}d$, seed 1 & 1.00 & 0.98 & 0.99 & 0.99 & 0.97 & 0.92 & 0.97 & 0.91 \\
\PAR-hold ($c{=}2$) + stretch FT, $R{=}288$ & 1.00 & 1.00 & 1.00 & 1.00 & 1.00 & 1.00 & 1.00 & 0.99 \\
Same FT without stretch & 0.98 & 1.00 & 0.91 & 0.69 & 0.23 & 0.26 & -- & -- \\
\bottomrule
\end{tabular}
\caption{$S_5$ accuracy by test depth (single-query proportions, chance 0.2). These earlier runs sample $R\le32$ in
training; the $c=1$, $R=d$ rows use $16\times$ the largest training loop count at $d=512$. Positional stretch helps only together with
progress-aligned targets.}
\label{tab:s5depth}
\end{table*}

\section{Additional Results}
\label{app:extra}

\noindent\textbf{Per-depth $S_5$ results for the first recipe version.}
Table~\ref{tab:s5depth} gives accuracy by test depth for \PAR-hold and the baselines (train $d\le16$).

\noindent\textbf{\PAR-hold on the swap-$S_5$ protocol.}
Across 6 seeds evaluated on 1{,}024 sequences per length, \PAR-hold with positional stretch achieves $84.5\pm29.4\%$
at $8\times$ and $76.9\pm38.0\%$ at $16\times$, trailing forced-depth \PAR{} in both mean accuracy and cross-seed
consistency.

\noindent\textbf{Explicit CoT by length.}
On the swap-$S_5$ protocol, the explicit CoT baseline achieves perfect accuracy at the trained length (100\% at 32
moves) but collapses rapidly beyond it: full-state accuracy drops to 35.6\% at 64 moves, 3.1\% at 128 moves, and
collapses to 0.8\% at 256 and 512 moves. Similarly, on the single-query $S_5$ task (trained on depth $\le16$), CoT
accuracy falls from 100\% at depth 16 to 12\% at 32, 4\% at 64, and 0\% at 128.

\noindent\textbf{Per-seed halting results.}
Table~\ref{tab:haltseeds} lists the per-seed results behind Tables~\ref{tab:halt} and~\ref{tab:haltrules}.

\begin{table}[]
\centering
\small
\setlength{\tabcolsep}{2.6pt}
\begin{tabular}{@{}lcccccc@{}}
\toprule
 & \multicolumn{3}{c}{256 ($8\times$)} & \multicolumn{3}{c}{512 ($16\times$)} \\
\cmidrule(lr){2-4}\cmidrule(lr){5-7}
Seed & $R{=}L$ & 6 of 8 & 8 in a row & $R{=}L$ & 6 of 8 & 8 in a row \\
\midrule
0 & 100.0 & 100.0 & 100.0 & 99.6 & 99.6 & 99.6 \\
1 & 99.9 & 99.9 & 99.9 & 99.9 & 98.3 & 99.9 \\
2 & 92.4 & 92.4 & 92.4 & 59.9 & 59.9 & 59.9 \\
3 & 100.0 & 100.0 & 100.0 & 99.9 & 99.9 & 99.9 \\
4 & 99.9 & 99.9 & 99.9 & 99.7 & 99.7 & 82.4 \\
5 & 100.0 & 100.0 & 100.0 & 99.9 & 99.9 & 99.9 \\
\midrule
mean & 98.7 & 98.7 & 98.7 & 93.2 & 92.9 & 90.3 \\
\bottomrule
\end{tabular}
\caption{Learned halting per seed (swap-$S_5$, full-state \%, 1{,}024 sequences per length, step-6000 checkpoints).
$R{=}L$: oracle iteration count. ``6 of 8'': select the first answer iteration $R$ with $P(\texttt{Yes})>0.5$ at which at
least 6 of the iterations $R..R{+}7$ exceed 0.5, then return the prediction from $R$ after confirmation;
``8 in a row'': all 8 must exceed 0.5.}
\label{tab:haltseeds}
\end{table}

\noindent\textbf{Per-run label-free results on swap-$S_5$.}
Table~\ref{tab:sdruns} reports prefix self-distillation results on the swap-$S_5$ protocol, before and after the
positional-stretch fine-tune.

\begin{table}[]
\centering
\small
\setlength{\tabcolsep}{3.5pt}
\begin{tabular}{@{}lccc@{}}
\toprule
Starting model & $8\times$ & $16\times$ & q-acc $16\times$ \\
\midrule
$n_k{=}2$, seed 0 & 73.8 & 0.4 & 22.1 \\
\quad + positional stretch & 76.2 & 26.7 & 74.9 \\
$n_k{=}4$, seed 2 & 52.7 & 0.0 & 26.7 \\
\quad + stretch & 99.9 & 99.7 & 99.8 \\
\quad + stretch, second FT seed & 99.2 & 98.4 & 99.2 \\
$n_k{=}4$, seed 4 & 72.3 & 5.5 & 39.4 \\
\quad + stretch & \textbf{100.0} & 99.7 & 99.9 \\
$n_k{=}4$, seed 5 & 30.9 & 1.2 & 32.7 \\
\quad + stretch & 86.2 & 65.5 & 85.2 \\
\midrule
Stretch from the start, seed 7 & \textbf{100.0} & \textbf{100.0} & \textbf{100.0} \\
\bottomrule
\end{tabular}
\caption{Prefix self-distillation on swap-$S_5$ (train $L\le32$, full-state \%; q-acc: per-query accuracy, chance
20\%). Models before the stretch fine-tune use 256 sequences per length, all others 1{,}024; every stretch fine-tuned
model scores at least 99.2\% in range (32 moves). The seed-4 and seed-5 starting models are step-5000 checkpoints.}
\label{tab:sdruns}
\end{table}

\noindent\textbf{Compute.}
All runs train in bf16 with DeepSpeed ZeRO-2 \citep{rajbhandari2020zero}, on two RTX 4090 or 3090 GPUs for
Qwen2.5-0.5B and Qwen3-0.6B and on eight H200 GPUs for Qwen2.5-1.5B, 3B, 7B, and Qwen3-1.7B and 4B.

\noindent\textbf{Configurations of the halting and label-free runs.}
Learned halting is trained jointly from the pretrained parent for one epoch with weight 0.5. The fine-tuned variant
starts from two \PAR{} checkpoints and trains for 0.3 epochs with learning rate $10^{-5}$. The label-free models use
forced depth, an EMA teacher with decay 0.99 read at its own completion iteration, and $n_k\in\{2,4\}$ distilled
iterations per example, trained for 0.6 and 1 epoch. The positional-stretch fine-tune continues each starting model
for 0.5 epochs with the same loss and $n_k$ (step-6000 checkpoint for seed 2, step 5000 for seeds 4 and 5, final
model for the $n_k{=}2$ seed 0). The stretch-from-the-start models train for one epoch with $n_k{=}2$.

\end{document}